# Izhikevich-Inspired Optoelectronic Neurons with Excitatory and Inhibitory Inputs for Energy-Efficient Photonic Spiking Neural Networks

YUN-JHU LEE, MEHMET BERKAY ON, XIAN XIAO, ROBERTO PROIETTI, AND S. J. BEN YOO[1*]

[1]*Department of Electrical and Computer Engineering, University of California, Davis, CA 95616, USA*
*\*Corresponding author: sbyoo@ucdavis.edu*

**Abstract:** Neuromorphic computing overcomes limitations of traditional von Neumann computing architectures by incorporating synaptic interconnections of neurons with nonlinear activation functions in a distributed neural network architecture. While some electronic neuromorphic computing demonstrations have shown significant energy efficiency improvements over their von Neumann counterparts, their scalability and throughput are still limited due to the impedance of the electrical wire interconnects. Photonic spiking neural networks (PSNNs) overcome the limitations of the electrical interconnects and potentially offer exceptionally high throughput and energy-efficiency compared to electronic neuromorphic counterparts while maintaining their benefits from event-driven computing capability. In this paper, we designed, prototyped, and experimentally demonstrated, for the first time to our knowledge, an optoelectronic spiking neuron inspired by the Izhikevich model incorporating both excitatory and inhibitory optical spiking inputs and producing optical spiking outputs accordingly. The optoelectronic neurons consist of three transistors acting as electrical spiking circuits, a vertical-cavity surface-emitting laser (VCSEL) for optical spiking outputs, and two photodetectors for excitatory and inhibitory optical spiking inputs. Additional inclusion of capacitors and resistors complete the Izhikevich-inspired optoelectronic neurons, which receive excitatory and inhibitory optical spikes as inputs from other optoelectronic neurons. We developed a detailed optoelectronic neuron model in Verilog-A and simulated the circuit-level operation of various cases with excitatory input and inhibitory input signals. The experimental results closely resemble the simulated results and demonstrate how the excitatory inputs trigger the optical spiking outputs while the inhibitory inputs suppress the outputs. Utilizing the simulated neuron model, we conducted simulations using fully connected (FC) and convolutional neural networks (CNN). The simulation results using MNIST handwritten digits recognition show 90% accuracy on unsupervised learning and 97% accuracy on a supervised modified FC neural network. We further designed a nanoscale optoelectronic neuron utilizing quantum impedance conversion where a 200 aJ/spike input can trigger the output from on-chip nanolasers with 10 fJ/spike. The nanoscale neuron can support a fanout of ~80 or overcome 19 dB excess optical loss while running at 10 GSpikes/second in the neural network, which corresponds to 100× throughput and 1000× energy-efficiency improvement compared to state-of-art electrical neuromorphic hardware such as Loihi and NeuroGrid.

## 1. Introduction

Machine Learning (ML) and Artificial Intelligence (AI) have already transformed our everyday lives--everything from scientific computing to shopping and entertainment involves some form of machine learning or intelligent algorithms. Such AI and ML systems typically use large von Neumann computing systems such as data centers, and they have shown remarkable capabilities to beat the human brain in some tasks, including the highly complex game of Go [1,2]. However, today's data centers consume megawatts of power (Google's AlphaGo utilized 1920 CPUs and 280 GPUs), and the current deep neural network algorithms require labor-intensive hand labeling of large datasets. While such high power consumption of cloud computing is currently tolerated due to the consolidated and amortized economic model of massive users, the newly emerging edge-computing [3] in the autonomous vehicles, drones, robots, smart-health, and the gateways of the Internet-of-Things (IoT) drives the need of intelligent, power-efficient, and high-throughput neuromorphic computing. For instance, future autonomous vehicles equipped with five LiDARs will require real-time information processing at 100 TOPS ($10^{14}$ Inference Operations per Second) at less than 100 Watt power consumption [4] with < 1 ms latency without relying on network connectivity to the cloud. Instead of using artificial neural networks (ANNs) in von Neumann architectures (e.g., utilizing non-spiking neural networks in GPU-based cluster), recent efforts towards spiking neuromorphic computing such as IBM's TrueNorth [5] and Intel's Loihi [6] processors have demonstrated significant energy-efficiency improvements compared to the non-spiking ANN counterparts. The bio-inspired neuromorphic hardware systems such as IBM's TrueNorth claim to achieve 176,000 times higher energy efficiency than the general-purpose Intel i7 based von-Neumann computing system for specific applications [5]. However, such electronic solutions typically include long electrical wires with large capacitance values leading to high interconnect energy consumption and requiring several repeaters for multi-hop connections to other non-neighboring nodes. For instance, the TrueNorth chip runs at slow speeds (i.e., 1 kHz), communicates with an energy efficiency of 2.3 pJ/bit (with an additional 3 pJ/bit for every cm of transmission), and requires a $256 \times 256$ cross-bar that selectively connects incoming neural spike events to outgoing neurons. The recently emerging nanoelectronic neuromorphic computing systems also suffer from similar communication challenges in achieving appreciable repeaterless distances, especially at high speeds [7].

Compared to electrical neural networks, photonic neural networks (PNNs) can significantly enhance energy efficiency and throughput by exploiting silicon photonic interconnects. On the other hand, event-driven spiking neural networks (SNNs) outperform artificial neural networks (ANNs) in terms of energy efficiency while keeping the same accuracy. Thus, it is beneficial to develop photonic spiking neural networks (PSNNs), which can take both the advantages mentioned above. Previous researches [8–10] exploited VCSEL's nonlinearity and vertical-cavity semiconductor optical amplifier(VCSOA) as a spiking neuron activation function on PSNNs. However, VCSEL neurons require a continuous power supply to maintain neuron behavior, therefore, violate the energy efficiency benefits. The other study [11] uses phase change materials (PCM) cells as spiking neurons and applies wavelength division multiplexing (WDM) techniques to sum up weights on synapses to the neuron. WDM-based PSNNs will have scalability problems due to the number of wavelength channels to operate when applying deep neural networks with thousands of neurons in multiple hidden layers. Besides, there is no energy benchmark on these photonic spiking neurons or PSNN compared to other electrical counterparts.

In a 2017 article [12], Miller reviews the possibilities of attojoule photonics and presents practical ~ 10 fJ/bit interconnect solutions with ~19 dB (80 ×) link loss budget exploiting quantum impedance conversion [13]. In this case, the signal transmission is not subject to the charging of capacitances but instead exploits close integration with electronics with less than 1fF capacitance. Hence, it is possible to realize nanophotonic devices closely integrated with nanoelectronics to form a neuron at 10 fJ/bit energy efficiency with a fanout of 10-100. When using low-loss waveguides, the neuron is capable of communicating with other neurons nearly independently of the communication distance at high speeds (> 10 GHz). These nanoscale optoelectronic neurons can be far more energy-efficient than all-optical non-linear neurons, which may require high optical energy per spike beyond 10 pJ/bit [11] and may suffer from challenges of isolating the input spikes while allowing output spikes [14]. That is why we first present our optoelectronic neuron design in this work. The optoelectronic neuron is designed with only three transistors, significantly reducing the transistor number compared to state-of-art design [15]. The transistors in the neuron control its behavior, and fewer transistors mean less power required to trigger the neuron. The neuron model was inspired by the Izhikevich model, which preserved biological neuron behavior and introduced three equations that govern the optoelectronic neurons. Besides, the optoelectronic neurons are designed to take excitatory and inhibitory inputs at the same time. This design makes our neurons resemble real biological neurons and offers new possibilities to construct large neural networks when interconnecting many of such neurons through photonic synaptic interconnections.

Recent progress in photonic matrix multipliers and arbitrary photonic couplers [16] indicates that photonic synaptic interconnects with arbitrary weight values can be trained without having electronic transistors repeated in the signal path. We can form a hierarchical synaptic interconnection with changing weight values through 2D and 3D photonic integrated circuits with reconfigurable photonic interconnection fabrics aided by nanoMEMS (NEMS) that can be remembered by latching NEMS component with little static energy consumption. Another possibility is to incorporate optical phase change materials (OPCM) such as GeSbTe (GST) [17] or GeSbSeTe (GSST) [18] in the optical synapse. As a result, the PSNN consisting of the attojoule nanoscale optoelectronic neurons and NEMS or OPCM synaptic interconnects can offer an extremely energy-efficient, high-throughput, and event-driven neuromorphic computing platform with near-zero static energy consumption for future edge computing.

In this paper, we discuss the design, simulation, and experimental demonstration of bio-inspired optoelectronic neurons. We describe future nanoscale optoelectronic neurons, introduce synaptic optical interconnect neural networks, and benchmark PSNNs consisting of the experimental bio-inspired optoelectronic neurons as well as the nanoscale optoelectronic neurons of the future. We show the successful training and inference of the PSNNs through simulations to achieve handwriting recognition utilizing MNIST datasets. Finally, we benchmark our scalable PSNN architectures exploiting the nanoscale optoelectronic neurons to provide benchmarking results. The result shows that our PSNN has better energy efficiency compares to other neuromorphic counterparts.

## 2. Bio-inspired Opto-Electronic Neuron Model Design

Our first task is to design, simulate, and prototype bio-inspired photonic spiking neurons. Many established models represent biological neuron behaviors. The Leaky-Integrate and fire (LIF) model [19], the Hodgkin-Huxley model [19], and the Izhikevich model [20,21] are among the most studied for neural networks simulations. In particular, previous studies of photonic spiking neurons [9,22] have primarily relied on the LIF model due to its simplicity. While the LIF model employs linear differential equations, the Hodgkin-Huxley model utilizes four ordinary differential equations for four state variables. The Izhikevich model introduces two partial-differential equations [20] to model most biological neurons in mammals' nervous systems effectively. Although the LIF model is easier to realize on electronic hardware, the LIF model's refractory part is not easily realizable on analog hardware. The Hodgkin-Huxley model can most closely resemble biological neurons; however, it is extremely computationally complex. In this paper, we will utilize the Izhikevich model to simulate and design the bio-inspired optoelectronic neurons and the spiking photonic neural networks. The governing equations of the Izhikevich model are as follows (*a*, *b*, *c,* and *d* are constant parameters):

$$\frac{dv}{dt} = 0.04v^2 + 5v + 140 - u + I \tag{1}$$

$$\frac{du}{dt} = a(bv - u) \tag{2}$$

$$when\ v \geq V_{threshold} \begin{cases} v \leftarrow c \\ u \leftarrow u + d \end{cases} \tag{3}$$

While the quadratic term of Equation (1) in the Izhikevich model helps produce results closely resembling biologically-observed spikes, it might cause instabilities in signals from equivalent optoelectronic circuits emulating the simulated neurons. Inspired by the Izhikevich model, we designed a spiking optoelectronic circuit neuron with excitatory and inhibitory inputs (see Fig. 1). The optical spiking inputs at the excitatory photodetector PD1_exc (inhibitory photodetector, PD2_inh) generate electrical current spikes which may (may not) drive the current into the output laser (a VCSEL) to generate (not generate) optical output spikes following the Izhikevich model. We introduce the following three equations (4)-(6) that govern the optoelectronic neurons of Fig. 1.

The membrane potential of the neuron is implemented as:
$$R_1 C_1 \frac{dv}{dt} = R_1(I_{exc} - I_{inh}) - R_1 K_1 \max\{0, u - V_{th1}\}^2 - v \qquad (4)$$

While the refractory potential can be realized as:
$$R_2 C_2 \frac{du}{dt} = R_2 K_3 max\{0, v - V_{th3} - u\}^2 - u \qquad (5)$$

A directly modulated laser implements the output of the optoelectronic neuron. We utilize commercial O-band Vertical Cavity Surface Emitting Laser (VCSEL) diodes in the testbed prototyping experiments. The amplitude of the VCSEL output signal is determined by $I_{VCSEL}$, which can be approximated as:
$$I_{VCSEL} = K_2 \max\{0, v - V_{th2}\}^2 \qquad (6)$$

$R_1$, $R_2$, $C_1$, and $C_2$ are the values of the resistors and the capacitors shown in Fig. 1. $K_1$, $K_2$ and $K_3$ are transconductance gain of the field-effect transistor (FETs). The model neglects subthreshold behaviors of the transistors.

Fig. 1 conceptualizes the working principle of the optoelectronic neuron. All transistors' operating points are set to the saturation condition. The optical spiking inputs detected by the excitatory photodetector *PD1_exc* will generate photocurrents to be integrated by the capacitor *C1* in the Membrane Potential Circuit (MPC) and discharged through the resistor *R1*. As the voltage (the Membrane Potential) of the Membrane Potential Circuit build-up to the threshold of FET1 and FET2 to drive current through VCSEL, it will fire output spikes, and the capacitance C2 in the Refractory Potential Circuit starts to charge up. As the refractory potential builds up to the threshold of FET3, the Membrane Potential is reset and kept at the reset voltage level until the refractory potential discharges below the threshold of FET3. More details about choosing parameters are provided in the supplementary information.

In addition to the excitatory connection, a critical feature included in the designed neuron model is the inhibitory connection. The inhibitory connection can be interpreted as a negative input to the neuron [23]. The inhibitory connection design is also presented in other neuron design [24], even if those designs focus on the ANN architecture instead of the SNN architecture. Inhibitory input signals decrease the accumulated membrane potential and make the neuron less responsive to the excitatory inputs. Contrary to the perceptron's negative inputs in ANNs, inhibitory inputs do not affect the neuron behaviors when the neurons are at the resting state. They are only effective when the neurons are excited beforehand. In the opto-electronic neuron hardware, excitatory and inhibitory inputs are received by the balanced photodetector (PD1_exc and PD2_inh) biased at the voltage $V_d$ in reference to the ground. Therefore, inhibitory inputs are ineffective when the opto-electronic neuron is at its resting state ($v(t) = 0$, where $v(t)$ is measured at the Membrane Potential).

The following Section 3 provides detailed simulation results of our optoelectronic neurons' governing equations and demonstrates a close match between the circuit-level simulation results and experimentally measured results (despite neglecting the subthreshold behaviors in the simulated model).

There are two main differences between the original Izhikevich model and the presented optoelectronic neuron model. The first is the quadratic positive feedback part of the membrane potential in the Izhikevich model, which is ignored due to the instabilities in the circuit when the signal is large. Here, the Izhikevich model considers the spike output solely as $v(t)$. In contrast, for optoelectronic neurons, another circuit stage involving FET2 and VCSEL in Fig. 1 is required to map the Membrane Potential voltage to drive electrical current into the VCSEL. The second difference lies in its threshold behavior, as shown in Equation (3). When the membrane potential reaches the threshold voltage, both variables, $v(t)$ and $u(t)$ are immediately set to their reset values. However, this mechanism implemented in the circuit behaviors involving FET3, R2, and C2 in the optoelectronic neuron will gradually reset in time, unlike in the Izhikevich model.

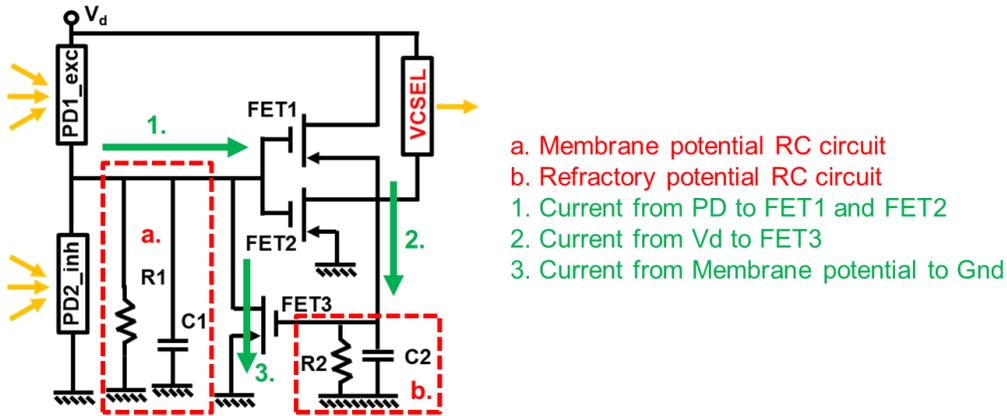

Fig. 1. Optoelectronic neuron design consisting of three transistors (FET1, FET2, and FET3), one optical output (VCSEL), and two optical inputs detected by photodetectors (PD1_exc for excitatory input and PD2_inh for inhibitory input). (a) Membrane potential RC circuit. (b) Refractory potential RC circuit. The current flow 1. Current from PD building membrane potential on a. 2. Generating output at VCSEL when a. reaching the threshold. 3. Refractory feedback control mechanism to drain current.

## 3. Opto-Electronic Neuron Simulation Results

To pursue bio-inspired optoelectronic neuron designs inspired by Izhikevich's model and to simulate the neuron behaviors, we built a two-level model consisting of the neuron circuit-level and the neural network-level models.

The circuit-level simulations focus on neuron circuit behavior. As Fig. 2 illustrates, we built a compact model on Verilog-A for the optoelectronic neuron. Here, we included details of the physical parameters, such as the transistor model. We simulated the model using LTSpice and Cadence to emulate the optoelectronic spiking neuron behaviors under continuous and discrete arbitrary spiking patterns. As an initial simulation example, we set the neuron behavior parameters to accumulate three continuous input spikes to cause the membrane potential to rise above the threshold for firing an output spike. We can easily change this neuron behavior by adjusting the capacitor and resistor values in the circuit to meet other neural network requirements. Section 4 will show the simulation results in detail.

For the neural network-level simulations, we used the Nengo simulator [25], which provides an excellent platform for performing neural network simulations that include neuron models, spiking neural networks (SNNs) learning algorithms, and synaptic interconnect models. To test our neuron performance on a neural network for an actual application, we imported the equations (4)-(6) and built our neuron model in Nengo. In addition to these equations, variables $u$ and $v$ are clipped between '0' (the ground level), and '$V_d$' (the maximum voltage level) and the minimum spike width of the optoelectronic neuron is scaled up from 0.1 ns to 1ms to match with Nengo simulation platform.

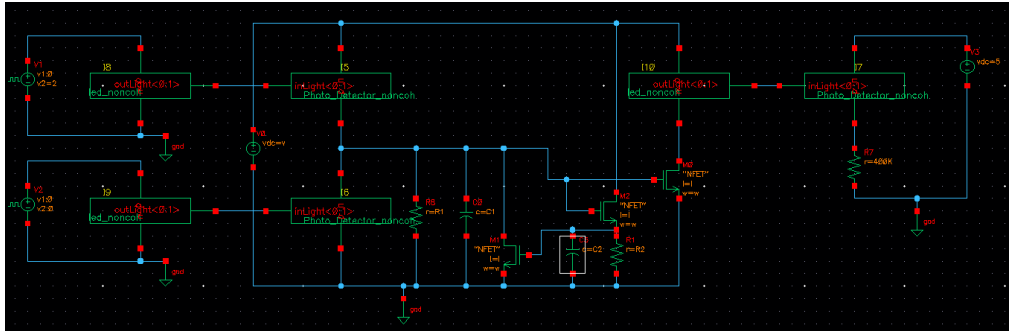

Fig. 2. Verilog-A neuron circuit model

Fig. 3 shows the simulated neuron behavior of our neuron model using different parameters. The simulated result in Fig. 3(a) matches nearly perfectly with experimental neuron results presented in Section 4. As expected from Izhikevich's model [20], various spiking patterns---fast, slow, and burst spiking neurons can be implemented by tuning the parameters in equations (4) and (5). The various parameters for emulating different neuron behaviors are actual physical parameters to be implemented in the optoelectronic neuron circuit. To verify neuron responses in simulations and experiments, we used an input spike train with four groups of spikes with a maximum spiking rate of 1 kHz. The spiking rate in the simulation is limited by Nengo simulator's minimum time step, which we can achieve a higher spiking rate by scaling down parameters in the experiment. The number of spikes in each group is 14, 5, 3, and 1 for the 1st, 2nd, 3rd, and 4th spike groups, respectively. There is a guard time of 30 ms between the groups. This guard time is to demonstrate that the discharge of the accumulated photocurrent shuts off the spiking output.

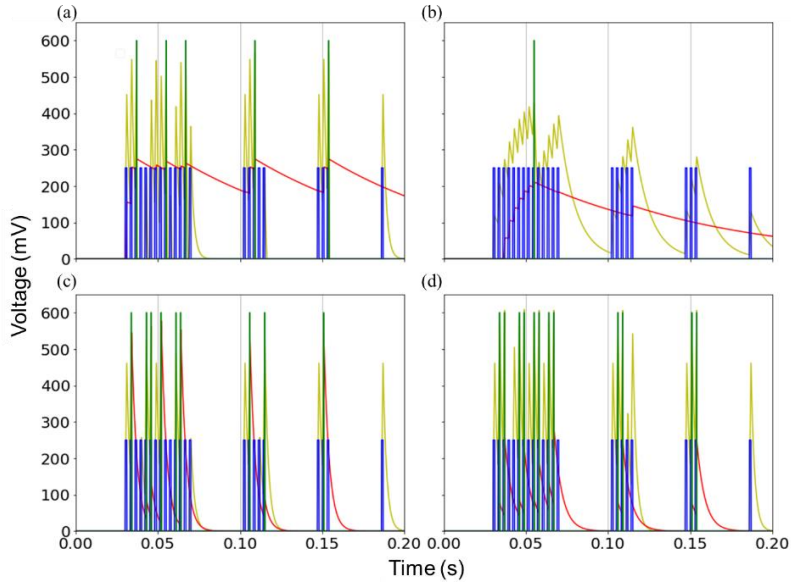

Fig. 3. Spiking neuron simulation performance. (a) is the regular spiking, which we will experimentally demonstrate in this paper. (b)(c)(d) is the neuron behaviors of our neuron model with different parameters. (b) is low output spiking, (c) is burst-spiking, (d) is fast output spiking simulation results. (Blue: excitatory input, Red: refractory potential, Yellow: membrane potential, Green: neuron output)

## 4. Opto-Electronic Neuron Experimental Results

Fig. 4 illustrates an experimental setup for our proof-of-principle neuron utilizing a commercial laser, commercial photodetectors PD1_exc and PD2_inh, and an electronic neuron circuit formed by discrete electronic components on a printed circuit board. To test this neuron's optical input and output spiking performance, the apparatus includes an FPGA generating arbitrary spiking patterns modulating the excitatory and the inhibitory lasers, Laser1 and Laser2, respectively. The outputs from Laser1 and Laser2 emulate the signals from upstream neurons. These outputs are directly coupled to PD1_exc (excitatory photodetector) and PD2_inh (inhibitory photodetector). For this first experimental demonstration, the FPGA output drives commercially packaged pigtailed lasers (Laser1 and Laser 2 in Fig. 4), while the output of the neuron circuit utilizes a 1.3-micron wavelength VCSEL (prototype VCSEL from VERTILAS®). The output spikes from the VCSEL are recorded by a Lightwave converter (Agilent Model #11982A) with optical input ports.

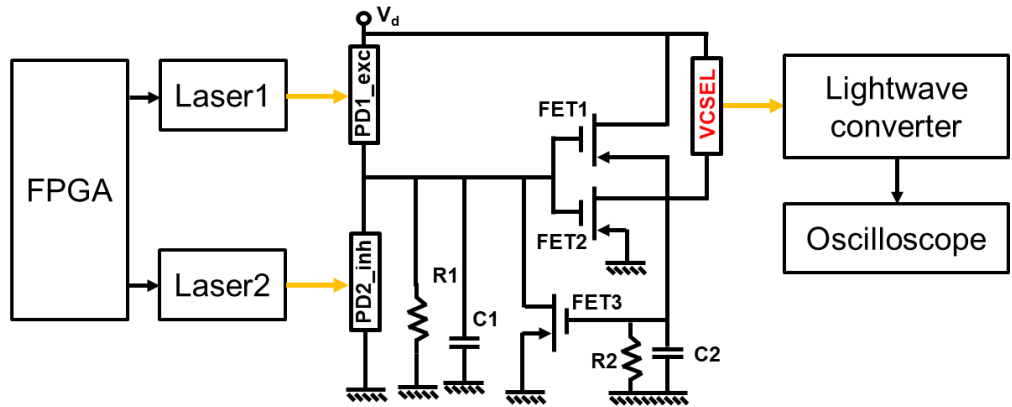

Fig. 4. An experimental setup for our proof-of-principle neuron utilizing a commercial laser, VCSEL, commercial photodetectors PD1_exc (excitatory photodetector) and PD2_inh (inhibitory photodetector), together with the electronic neuron circuit formed by discrete electronics on a printed circuit board. To test this neuron's optical input and output spiking performance, the apparatus includes an FPGA generating arbitrary spiking patterns into the excitatory and the inhibitory lasers, Laser1 and Laser2, respectively, directly coupling to PD1_exc and PD2_inh, respectively. An oscilloscope records the neuron's optical output of the laser.

Based on our circuit-level and neural network level simulation results utilizing the LTSpice® tool, we conducted comparative studies of simulation vs. actual hardware experimental results of our optoelectronic neuron behavior. For this purpose, we created four different spiking groups as mentioned in the previous section--- the number of spikes in each group is 14, 5, 3, and 1 for the $1^{st}$, $2^{nd}$, $3^{rd}$, and $4^{th}$ spike groups, respectively, with a guard time of 3 ms between the groups. This experiment is a version of the simulation described in Fig. 3. We scaled the parameters to surpass the minimum step limit on the Nengo simulator and allow experiments at a 10× faster time scale. Fig. 5 summarizes the results with the excitatory input signal only. Fig. 6 summarizes the results with both excitatory and inhibitory signal inputs.

Fig. 5(a) is the LTSpice® simulated input spiking pattern consisting of the four groups in sequence (14, 5, 3, and 1 spike in each group), and Fig. 5(b) is the measured optical spiking pattern output from Laser1. We observe some noise on the measured optical spikes consistent with the measurement setup. Fig. 5(c) and (d) provide the simulated and measured membrane potential values measured at the Membrane Potential Circuit of Fig. 1 by placing a monitor in the simulator and placing a probe in the actual experiment. As indicated by the arrows on Fig. 5(c) and (d), we observe that the simulated and the measured membrane potential values reach the threshold after three consecutive spike inputs. For the first spike group of 14 spikes, it reaches the threshold three times, and for both the second spike group of 5 spikes and the third spike group of 3 spikes, it reaches the threshold only once. For the fourth group of a single spike, it does not reach the threshold. Fig. 5(e) and (f) illustrate simulated and experimental results, including the refractory potential and the optical output from the laser in addition to the optical excitatory input and the membrane potential. Here we observe that the optical output spikes fire when the membrane potential reaches the threshold, but more importantly, the refractory potential rises in response to the spike output. This indicates that the membrane potential changes with optical input spikes and that the firing of the optical output spikes occurs only after the refractory period. This proves that our neuron model correctly represents the Izhikevich model. The experimental result in Fig. 5(f) shows the spike output behavior closely matching the LTSpice result in Fig. 5(e).

We repeated the simulation and the experiment when using both excitatory and inhibitory input signals by adding the inhibitory signal to PD2_inh. Fig. 6(a) and (b) show the additional inhibitory input signal (red) as marked by arrows (red) for the simulation and the experiment results. As Fig. 6(c) and (d) demonstrate, the neuron behaves similarly to the behaviors shown in Fig. 5(c) and (d) when the inhibitory signal is absent (the membrane potential rises to the threshold for three consecutive spikes as labeled as #1, #2, and #4 in Fig. 6(c) and (d)). When the inhibitory signals are present, the membrane potential gets frustrated and cannot accumulate charge to generate spikes at #3 and #5. This behavior contrasts with the behavior seen at #3 and #5 of Fig. 5(c) and (d), where the membrane potential rose to the threshold in the absence of the inhibitory signals. Fig. 6(e) and (f) are the LTSpice simulation and the actual experimental results of the neuron behavior, including the optical output spikes that fire only when the membrane potential reaches the threshold. The experimental results closely match the simulated results, and the optical output spikes are absent at #3 and #5 due to the presence of the inhibitory signal. The inhibitory signal (Red) cancels out the effect of the excitatory signal (Blue) to suppress the output spike (Green). This neuron behavior is consistent with the commonly seen functionality of biological inhibitory neurons.

## 5. Benchmarking Photonic Spiking Neural Networks

In this section, we benchmark optoelectronic neural networks resulting from the synaptic interconnection of our optoelectronic neurons against state-of-the-art electronic neural networks. This benchmarking section includes two parts. In the first part, we simulate the training and the inference capability of optoelectronic neural networks, including the designed optoelectronic neurons. Since the optoelectronic neurons serve as the neural network's activation function, the neuron's behavior will affect the inference and training accuracy. The second part of the benchmarking will address the energy efficiency and the throughput of the optoelectronic neural networks. In the benchmarking studies, we will consider the case of optoelectronic neurons used in our actual testbed experiments employing bulky commercial lasers and transistors. More importantly, the new case of nanoscale optoelectronic neurons with integrated nanotransistors (FETs compatible with 5 nm CMOS technologies). We consider three types of Izhikevich-inspired optoelectronic neurons. The first is the optoelectronic neuron described for our experimental testbed demonstration (Testbed neurons). The second is the foundry-implementation of optoelectronic neurons utilizing the same design process using the Izhikevich-inspired model but incorporating commercial silicon-CMOS-photonic foundry process (Foundry-neurons). The third is the nanoscale optoelectronic neurons (denoted as Nano-neurons) utilizing the quantum impedance conversion between nanoscale detectors and nanoscale (5nm) FET circuits driving nanoscale lasers (e.g., photonic crystal lasers). Further, for optical synaptic interconnections, we consider optical 2 × 2 Mach-Zehnder interferometric couplers in neural networks [26] [27] capable of achieving near-zero static energy consumption by incorporating optical phase shifters with optical MEMS devices [28] or with optical phase change materials [18,29–32].

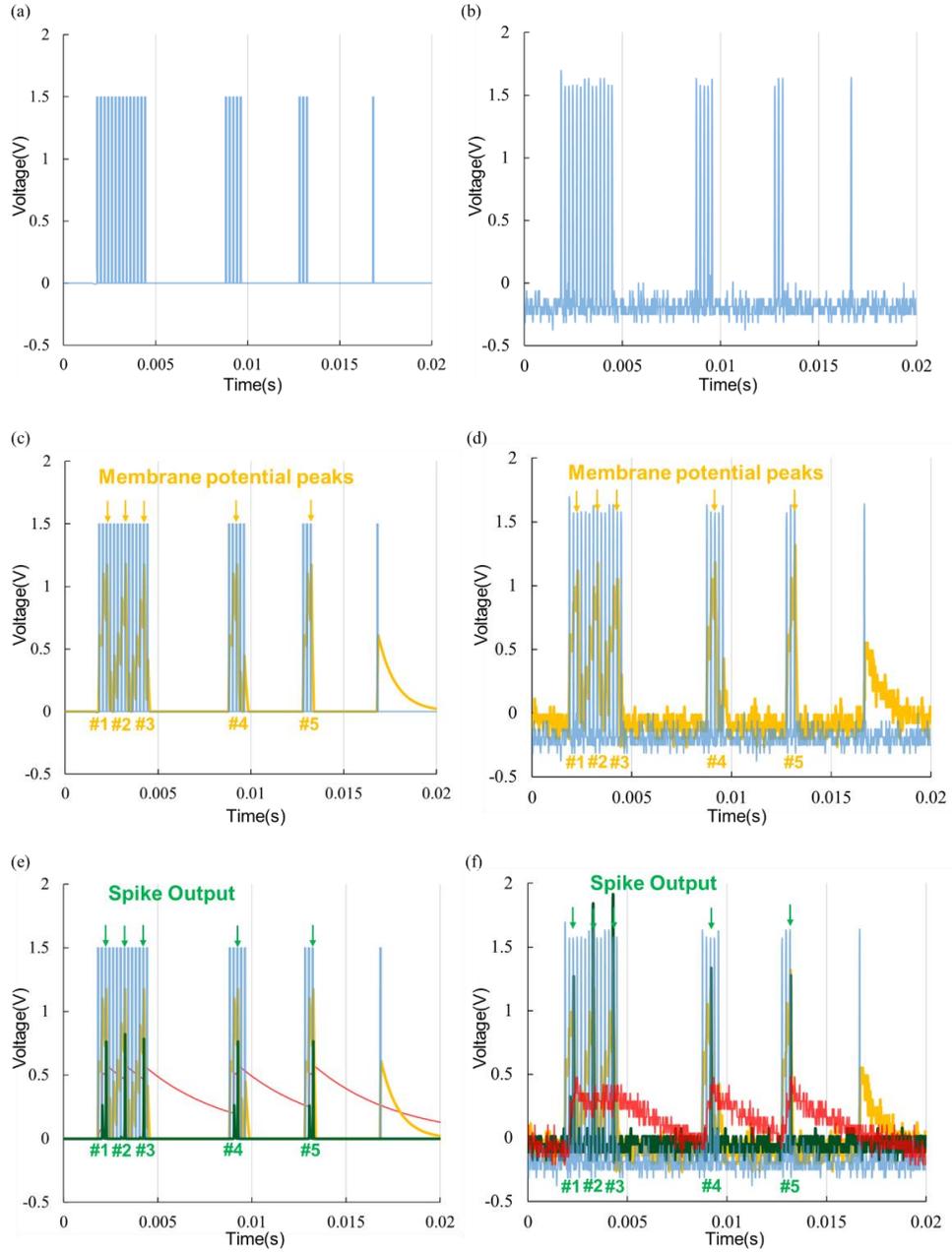

Fig. 5. Neuron spiking behavior with only excitatory signal input simulation and experiment. The input spiking pattern consisting of the four groups in sequence (14, 5, 3, and 1 spike in each group). (a) simulated inputs. (b) is the measured optical spiking pattern output from Laser1. We observe some noises on the measured optical spikes consistent with the measurement setup. (c) simulated membrane potential values. (d) provides measured membrane potential values measured at the Membrane Potential Circuit of Fig. 1 by placing a monitor in the simulator and by placing a probe in the actual experiment. (e) and (f) illustrate simulated and experimental results, including the refractory potential and the optical output from the laser in addition to the optical excitatory input and the membrane potential. The refractory potential rises in response to the spike output. This indicates that the membrane potential changes with optical input spikes and that the firing of the optical output spikes occurs only after the refractory period, which proves that our neuron model correctly represents the Izhikevich model. (Blue: optical excitatory input, Red: refractory potential, Yellow: membrane potential, Green: optical output)

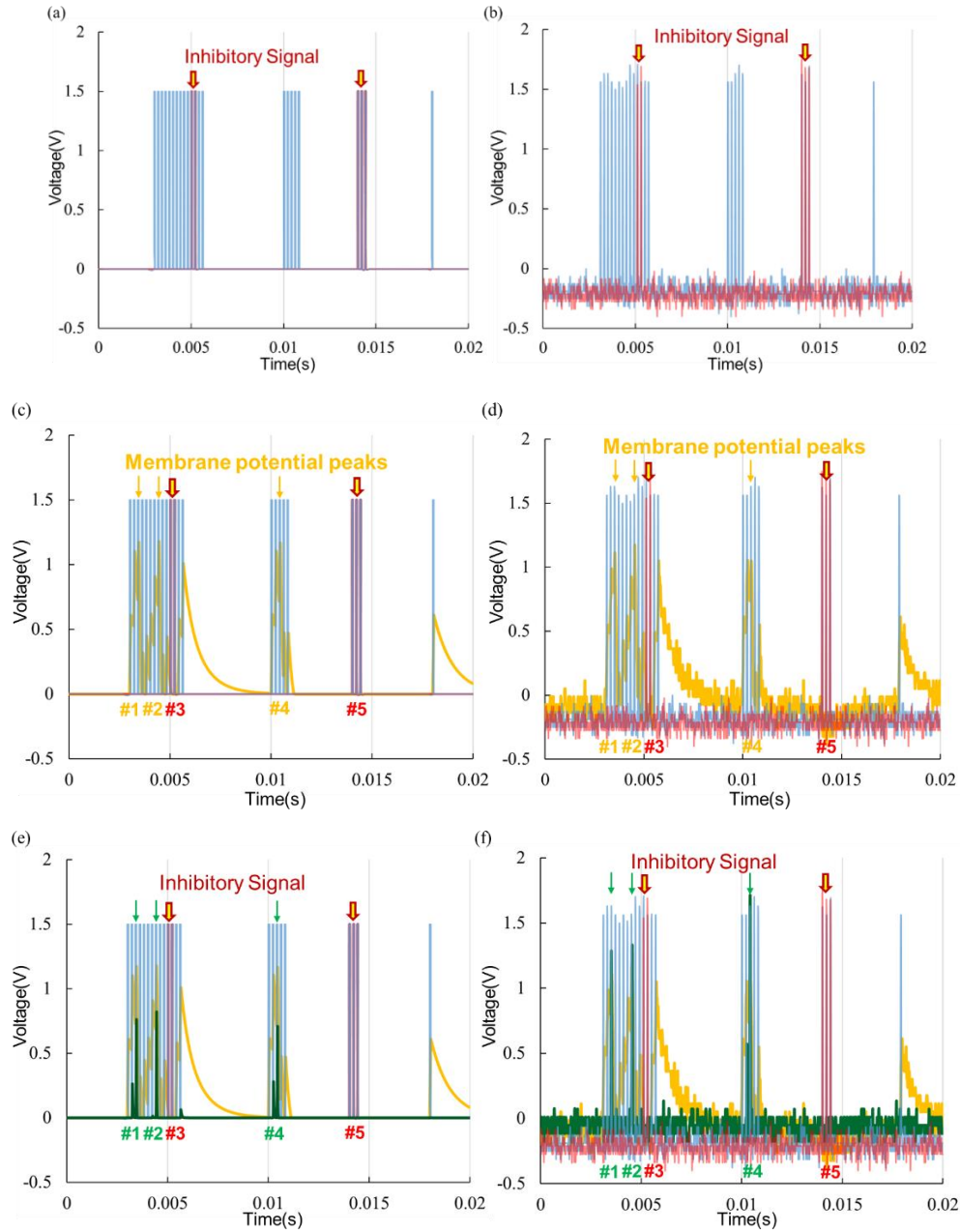

Fig. 6. Neuron spiking behavior with both excitatory and inhibitory signal inputs. The input spiking pattern for excitatory input is the same as Fig. 5, consisting of the four groups in sequence (14, 5, 3, and 1 spike in each group). An additional inhibitory signal is added on PD2_inh. (a) simulated inputs. Inhibitory inputs are label in red. (b) is the measured optical spiking pattern output from Laser1.   (c) simulated membrane potential values. (d) provides measured membrane potential values measured at the Membrane Potential Circuit of Fig. 1 by placing a monitor in the simulator and by placing a probe in the actual experiment. When the inhibitory signal is absent (the membrane potential rises to the threshold for three consecutive spikes as labeled as #1, #2, and #4 in Fig. 6 (c) and (d)), but when the inhibitory signals are present, the membrane potential gets frustrated. It cannot accumulate at #3 and #5. (e) and (f) illustrate simulated and experimental results, including the refractory potential and the optical output from the laser in addition to the optical excitatory input and the membrane potential. The experimental results closely match the simulated results, and the optical output spikes are absent at #3 and #5 due to the presence of the inhibitory signal. (Blue: optical excitatory input, Red: optical inhibitory input, Yellow: membrane potential, Green: optical output)

Fig. 7 shows the neural network structure with the optoelectronic neurons and optical synaptic interconnects. We simulated two cases of optoelectronic neurons in Fig. 7. Fig. 7(a) is the conventional feedforward neural network structure with inhibitory and excitatory signal connections. Fig. 7(b) has one-to-one inhibitory signal connections in the hidden layer instead. The neuron design in Section 2 supports both conventional feedforward neural network structures with inhibitory and excitatory signal connection for Fig. 7(a) and one-to-one inhibitory signal connection for Fig. 7(b). The inhibitory signals are a crucial element to realize the neural network behavior in neuroscience. In the first structure, both inhibitory and excitatory signals are transmitted to the optical synapses. Whereas in the second case, only excitatory signals will transmit on the optical synapses since the inhibitory signals are directly transmitted by the inhibitory neuron. In this topology, there are the same numbers of excitatory and inhibitory neurons in the hidden layer. The hidden layer performs the winner-takes-all setting by connecting one excitatory neuron to other excitatory neurons and sending negative feedback signals to inhibit signal transmission to the output layer.

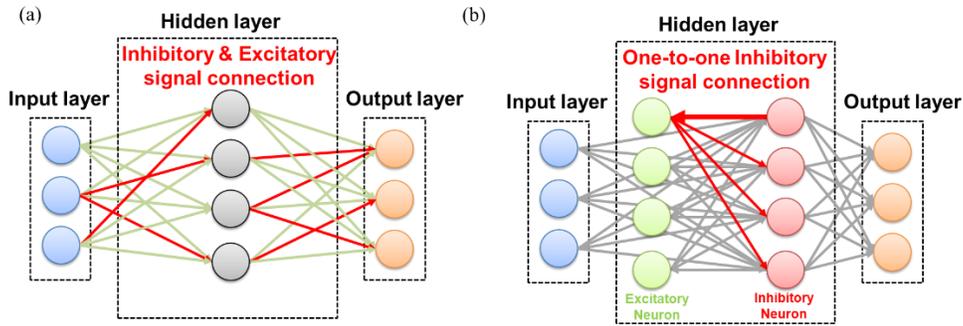

Fig. 7. Two neural network schemes. (a) is a neural network with inhibitory and excitatory signal coexist in the same interconnection (b) is a neural network with inhibitory and excitatory signals in separated interconnection. The inhibitory neuron only connects to one excitatory neuron (one-to-one inhibitory signal connection).

The synaptic interconnects between our optoelectronic neurons exploit the results from [33], which assign weight values by changing each mesh's phase shifter. Conventional multiport interferometers use electro-optical or thermal-optical tuning that will require a supply of constant power to maintain the phase states. However, integrating non-volatile optical MEMS [26] into our optical synaptic design will only require power when weight value changes on the neural network. This approach can significantly reduce the power consumption during training and perform zero power consumption on PSNNs during real task work.

The feedforward neural network structure with inhibitory and excitatory signal connection is benchmarked with FC Nets and ConvNets [34–36] neural network architectures. Fig. 7(a) shows the network architecture we used in Nengo and trained with supervised learning. Table 1 summarizes the inference benchmarking results with supervised learning. Here, we apply the ANN-to-SNN conversion training method. The FC Nets and ConvNets ANN are trained using the ReLU activation function with backpropagation methods. Next, we transferred the trained weight values to the proposed PSNN for testing. Table 1 compares the accuracy of the electronic neural network, the PSNN with LIF neuron, the PSNN with the proposed optoelectronic neuron, and the PSNN with future energy-efficient optoelectronic neuron. The PSNN with future energy-efficient optoelectronic neuron uses an abstract model to simulate on Nengo simulator. Our optoelectronic neuron model can reach 97% accuracy and achieves comparable results to LIF neurons.

For benchmarking of the one-to-one neural network of Fig. 7(b), we employed Diehl and Cook's MNIST handwriting recognition experiment [37] with 28 by 28 input neurons, 400 neurons in the hidden layer, and 10 neurons for output classification on our BRIAN simulator.

We used unsupervised learning Spike-timing-dependent plasticity (STDP) for training. We apply the winner-takes-all setting in the hidden layer. The excitatory neuron will send spike messages to its inhibitory neuron and provide feedback to suppress other excitatory neurons. By training with 60,000 MNIST datasets, we can achieve 90% accuracy [38] on this test (Table 2). We expect that fine-tuning the synaptic weight values and increasing the number of layers or neurons could improve the accuracy. However, this is out of the scope of this paper.

The benchmarking results show that our optoelectronic neuron can obtain accuracy similar to state-of-the-art LIF neuron-based neural networks on supervised and unsupervised learning neural network architecture. These results support that we can preserve more biological neuron behaviors while keeping the same accuracy of the conventional simplified neuron model.

Table 1. Neural network performance results on Nengo simulator based on supervised learning [34–36] with ANN to SNN conversion

| Network Type | Neuron Type | Artificial Neuron | LIF Spiking Neuron | Testbed Spiking Neuron | Foundry Spiking Neuron | Nano Spiking Neuron |
|---|---|---|---|---|---|---|
| FC Nets | 300-100-10 | 95.3 | 97.3 | 96.24 | 96.28 | 96.44 |
| FC Nets | 1000-500-10 | 99.51 | 97.55 | 97.04 | 97.12 | 97.55 |
| FC Nets | 1500-1000-500-10 | 99.54 | 98.03 | 90.72 | 91.36 | 92.31 |
| Conv Nets | LeNet-1 | 98.3 | 97.5 | 93.01 | 92 | 93.24 |
| Conv Nets | LeNet-5 | 99.05 | 98.03 | 83.59 | 84.49 | 90.95 |

Table 2. Neural network performance results on Brian simulator based on unsupervised spike-timing-dependent plasticity (STDP) learning [37]

| Network | Neuron Type | Diehl and Cook Test Accuracy | Optoelectronic Neuron Test Accuracy |
|---|---|---|---|
| 784-400-10 | | 95% | 90% |

We calculated the energy efficiency and the computing throughput of our neuromorphic computing platform compared to other electronic counterparts. As mentioned earlier, we considered both the neurons implemented in our current testbed, the foundry, and the future nanoscale optoelectronic neurons into the comparison to indicate the advantage of using PSNNs.

Our neural network's energy consumption occurs in two locations: the photonic MZI mesh for all-to-all synaptic interconnect and the neuron for non-linear function. Photonic MZI mesh networks have to consider dynamic and static power consumption. Conventional silicon photonic MZI mesh networks use thermo-optical-tuning of the phase-shifters, which typically consumes a continuous 10 mW power supply to keep the desired state of the phase-shift (~10 mW static power per MZI). Our testbed version power benchmark is based on the thermo-optically tuned MZI mesh. The static energy degrades the overall benchmark power-efficiency performance. However, for optical phase shifters consisting of silicon photonic MEMS [28], the weight values can be remembered by latching MEMS components with little static energy consumption. While the latching MEMS's reconfiguration is estimated to consume energy at ~1 pJ per reconfiguration, the impact on the energy efficiency is negligible since such reconfigurations are expected to be infrequent (below 0.1% duty cycle). Similar to optical MEMS-based MZI mesh, optical phase change materials (OPCM) such as GeSbTe (GST) [17] or GeSbSeTe (GSST) [18] can also achieve zero static energy synaptic interconnects. In contrast, dynamic energy consumption is required during the training phase of the neural network. Once the training process is completed, there is no additional power required to maintain the phase-shift states. In the following, we discuss benchmarking of energy efficiency, throughput, and accuracy of the neural networks available in the literature, our experimentally demonstrated testbed neural networks, our foundry implementation of the PSNN with MEMS MZIs (labeled as Foundry-PSNN), and our futuristic nanoscale optoelectronic neuron based PSNN with MEMS MZIs (labeled as Nano-PSNN).

Neuron's power consumption varies differently among devices. Current research, such as [12], usually requires 12-20 transistors to complete the neuron behavior, while our neuron structure only requires three transistors to perform neuron behavior. We can separate our neuron structure into three parts: photodetector, transistor circuit, and laser.

The testbed version uses an off-the-shelf photodetector, transistor, and laser. At the same time, the Foundry-PSNN consists of 90 nm CMOS with monolithically integrated silicon photonics, including MEMS phase shifters for MZI synaptic interconnects realized by 90 nm silicon photonic CMOS process. Thus Foundry-PSNN is a miniaturization of the current testbed-PSNN utilizing a commercial foundry with modified post-fabrication to realize MEMS MZI synaptic interconnects and micro-transfer-printed quantum dot lasers [39]. The detailed design of the Foundry-PSNN is shown in Fig. 8.

The nanoscale optoelectronic neuron exploits attojoule photonics [17] with quantum impedance conversion [18], where the signal transmission is not subjected to the charging of capacitance but rather exploits close integration with electronics with < 1 fF capacitance. The charging process on nanoelectronics includes load capacitance and the capacitor in the circuit. Our neuron's nanoelectronics includes the load capacitance on the photodetector, membrane capacitor, and transistor gate capacitance. The photodetector's load capacitance is around 0.1fF [3], and the simulated membrane capacitor is 0.5fF. The value of the transistor gate capacitance is derived from IRDS2020 [40]. Hence, it is possible to realize nanophotonic devices closely integrated with nanoelectronics to form a neuron at 10 fJ/bit energy efficiency with a fanout of 10-100. When using low-loss waveguides, the neuron is capable of communicating with other neurons nearly independently of the communication distance at high speeds (> 10 GHz).

Now, we will describe the design and analysis of our futuristic Nanoscale-PSNN. Fig. 9(a) shows the structural layout of the proposed nano-optoelectronic neuron. We propose a low-$Q$ nanophotonic crystal PD based on Ge/Si cavity. An ultra-low capacitance nano-cavity PD can generate sufficiently large voltage without an amplifier when combined with a high impedance load [13]. Based on this configuration, ~0.1fF capacitance is expected in the resonant nanophotonic PD. In addition to the ultra-compact size and extremely low capacitance, the extremely short electrical contact between PDs and next stage FET transistors can further guarantee extreme low circuit power consumption [12]. We anticipate such a system can operate beyond 10 GHz bandwidth with ultralow energy consumption of < 1 fJ/bit.

We will fabricate all devices on a silicon platform in which the FET, Ge/Si nanodetectors, and waveguides will be on silicon on silicon-oxide waveguides. At the same time, nanolasers will utilize hybrid InAs/AlGaAs quantum-dot on silicon/SiO$_2$ structure with photonic crystal patterns etched in on silicon. The hybrid InP Multi-Quantum-Well / silicon semiconductor optical amplifier demonstrated in [41] utilized a similar fabrication process. The absence of capacitive charge associated with the interconnect wires [12] can drastically reduce future nanophotonic neurons' power consumption. The transistor circuit's power consumption is determined by the number of transistors, the operation voltage, and transistor on-state current [42]. We set our neuron threshold voltage near the transistor's threshold to get the most energy efficiency. The output laser can be viewed as a diode in our neuron circuit. The expected nanolaser energy consumption is ~4.4 fJ [12]. The area of the device mentioned above is based on [12,43]. For our future version with 784-1000-500 neuron PSNN architecture for Diel-Cook implementations [37], we expect the chip area will be around 4 cm$^2$. By design of the nanophotonic and nanoscale FET structures, the static power consumption is negligible. Further, we expects to greatly improve the energy efficiency per spike event while achieving the desired neuron behaviors following the Izhikevich's model as testbed neurons did. Table 3 shows the energy and power calculation results. The detailed calculation is provided in the supplementary information.

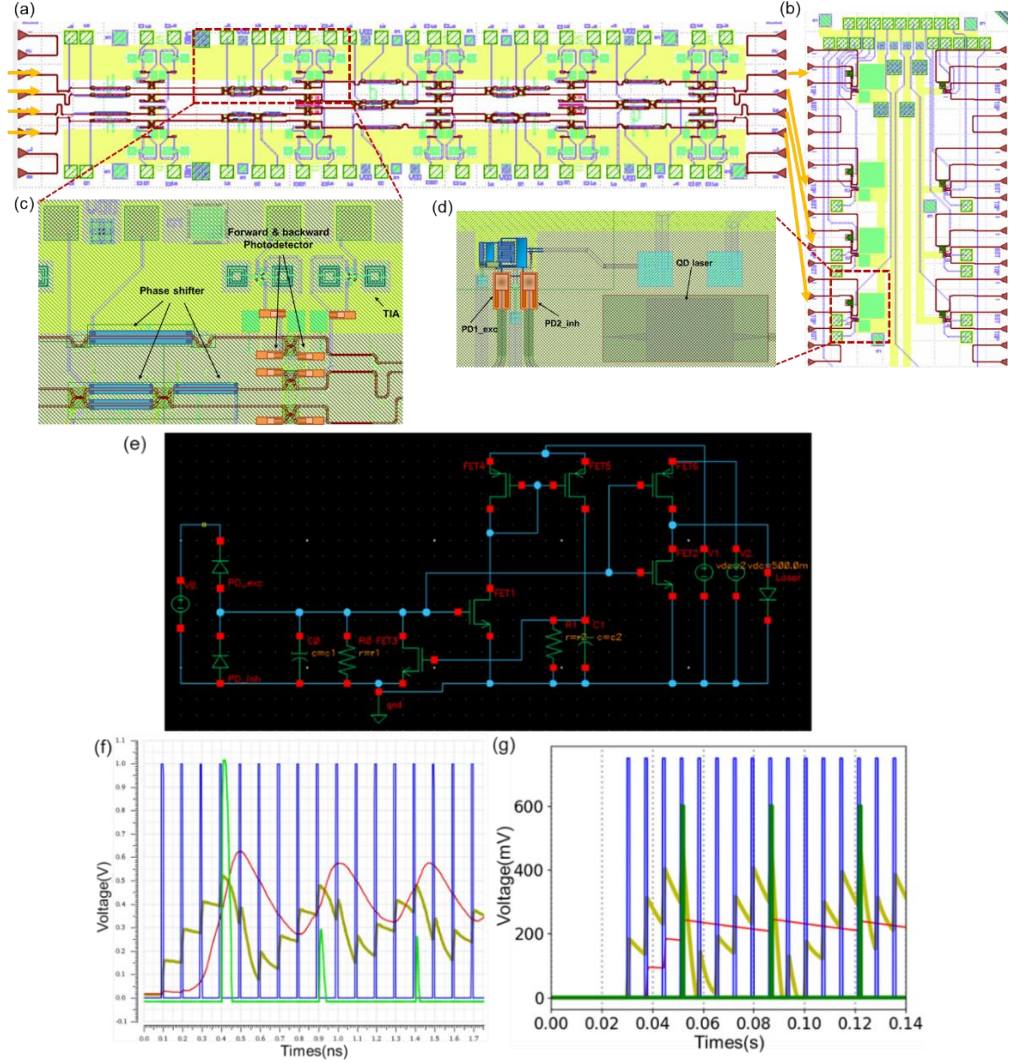

Fig. 8. The foundry-PSNN architecture consists of cascaded layers of a (a) MZI mesh synaptic interconnect network and (b) neuron layer. (a) is the 4 × 4 rectangular MZI mesh with embedded Bi-directional PD with a transimpedance amplifier (TIA) to support forward propagation and backpropagation training. (b) is a neuron chip with multiple neuron designs. There are 6 transistor model neurons with disk or ring modulated laser and neurons with micro-transfer-printed quantum dot (QD) lasers. The QD laser will fabricate separately. (c) is the detailed structure of 4×4 rectangular MZI mesh. The forward and backward PD is embedded for neural network training. (d) is one of our optoelectronic neuron designs used to connect micro-transfer-printed quantum dot lasers as neuron output. (e) is Spectre simulation of foundry version excitatory & inhibitory input coexist on neuron. (f) is the foundry version neuron cadence simulation at a 10GHz spiking rate. (blue: optical excitatory input; yellow: membrane potential; and green: laser modulation signals) (g) is the foundry version neuron spiking behavior used in Nengo simulation. The spiking rate in the simulation is limited by Nengo simulator's minimum time step. (blue: optical excitatory input; yellow: membrane potential; and green: laser modulation signals)

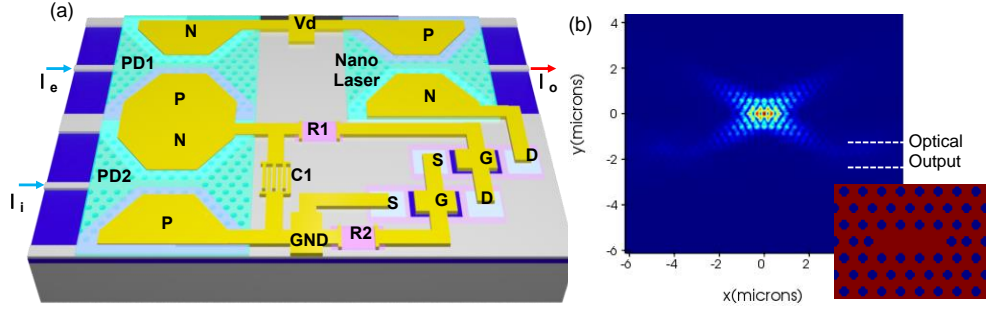

Fig. 9. (a) A schematic of the proposed optoelectronic neuron structure based on Fig. 1 including two Ge/Si photonic crystal enhanced photodiodes for excitatory and inhibitory inputs and two FETs on SOI for thresholding and spiking signal generation by triggering a photonic crystal cavity laser for in-plane emission. (b) a photonic crystal cavity laser will be fabricated on silicon utilizing heterogeneous integration by transfer-printing to realize hybrid III-V/silicon nanophotonic devices.

Table 3. Foundry and nano neuron energy and power consumption

|  | Foundry neuron | Nano neuron |
|---|---|---|
| $E_{dynamic-in}$ | 21.09fJ/spike | 200aJ/spike |
| $P_{static-off}$ | 6.36μW | 14nW |
| $P_{static-on}$ | 858μW | 43.78μW |
| MNIST handwriting recognition total training energy | 31.3μJ | 253nJ |

Fig. 10 shows an example of (a) the proposed Nano-Optoelectronic Neural Network (3 layer example) consisting of (b) Nanoscale-Optoelectronic Neurons, (c) self-optimizing [16] nanophotonic synaptic interconnect network with (d) 2×2 NEMS-MZI including tunable NEMS phase shifters. An essential part of a neural computation scheme is providing this necessary set of weighted connections from the nanoscale optoelectronic neuron outputs (lasers) and the inputs to the next layer of neurons (photodetectors).

Until now, there is no standard method or architecture for benchmarking spiking neural networks because while artificial neural networks are based on synchronous multiply-and-accumulate (MAC) operations, SNN computations are based on spike events. Thus, we considered three kinds of benchmarks that standardize the neural network approach for our PSNN. In the first one, we target spike-event-based computations, which all the competitors use SNN for image classification. We defined an operation (OP) as one spike event in the neural network in this first benchmark. One spike event starts from the neuron aggregating all the inputs from the previous layer, and it ends with generating an output for the next layer. The hardware perspective includes the energy consumption of synapses from previous layer neurons to the neuron, which generates a spike for the next layer. Using this approach, we eliminate the difference between classical von Neumann hardware and neuromorphic non-von Neumann hardware. To make the benchmarking fair between different hardware solutions, we only consider power consumption for inference in the neural network rather than training power consumption since different training algorithms and offline training systems for PSNN will result in different results.

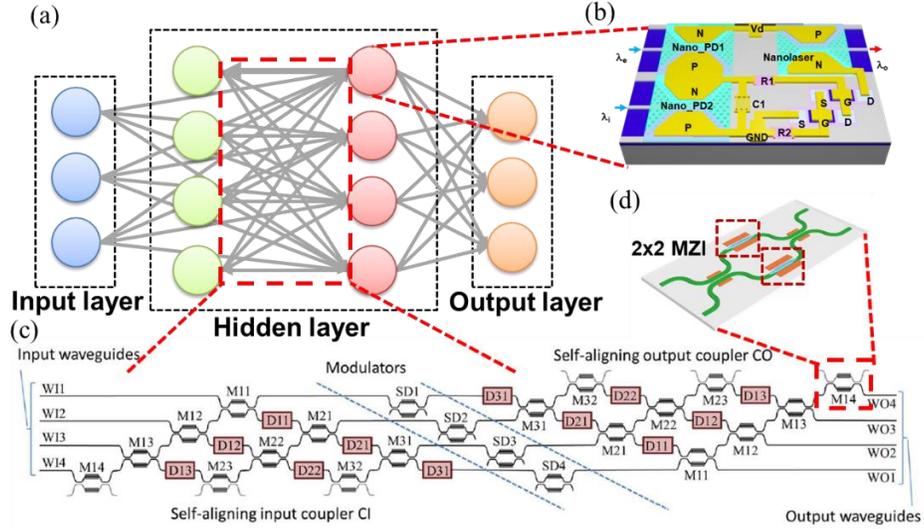

Fig. 10. (a) Proposed Nano-Optoelectronic Neural Network (3-layer example) consisting of (b) Nano-Optoelectronic Neurons, (c) self-optimizing [16] nanophotonic synaptic interconnect network with (d) 2x2 NEMS-MZI Synapse including tunable NEMS phase shifters or Phase Change Materials (PCM).

Fig. 11 is the plot of benchmarking results based on our first benchmarking method. Compared to state-of-art neuromorphic hardware [5,6,44–49], our proof-of-concept testbed version of PSNN can get around 0.001 GOP/s/W energy efficiency at 0.001OP/s/mm$^2$ computing speed. Our current Foundry-PSNN version can achieve around $5 \times 10^4$ GOP/s/W energy efficiency at 10 GHz spike-event speed. If we further utilize sub-10nm transistor and closer integration of nanophotonic and nanoelectronics, we can achieve over $10^6$ GOP/s/W energy efficiency at 10 GHz spike-event speed. We also list several neural network hardware that applying SNN in the plot. Our Foundry-PSNN version outperforms all state-of-art spiking neural network hardwares [5,6] by at least 1000× in terms of energy efficiency.

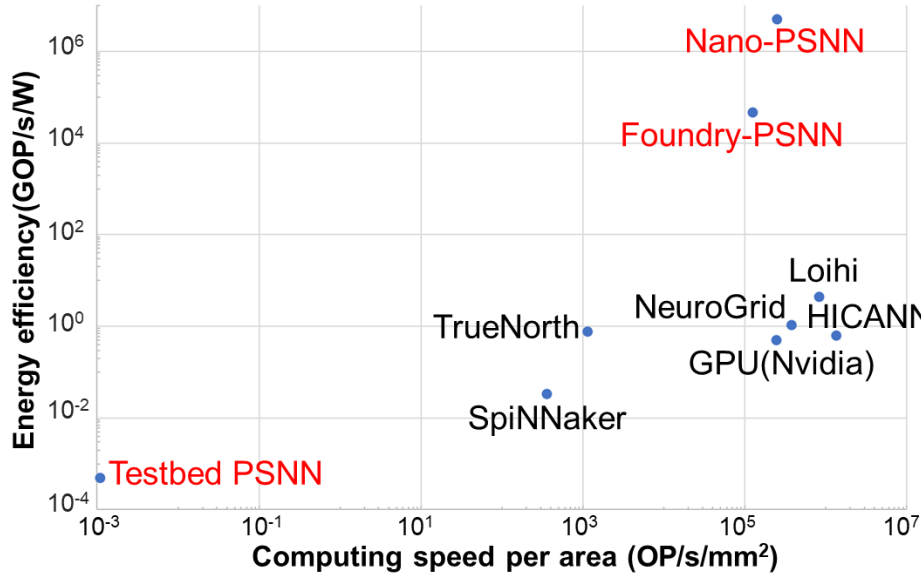

Fig. 11 Energy efficiency benchmarking method based on spike-event, including the PSNN, described in this paper (denoted in red) in comparison with results described in the literature ( [5,6,44–49]).

The second method of benchmarking is based on calculating the conventional multiply-accumulate (MAC) operation. In this comparison, we include both ANN and SNN to have a paramount view of energy efficiency between different approaches. One thing to notice here is that for SNN, the number of spikes per MAC would alter if we targeted a different task. Thus, the result shown here is based on applying PSNN on the MNIST dataset. We directly compare our PSNN with the energy efficiency results in [50]. The analog hardware is based on[5], [45], [47], [49]–[51]. The digital hardware is based on [54–58]. The photonic hardware estimation is based on [50]. As shown in Fig. 12, our GF-enabled version with O-E-O neural network can outperform most of all photonic approaches and achieve over $10^{26}$ [(MAC/s/mm$^2$)/(J/MAC)] energy efficiency. The future version can reach over $10^{29}$ [(MAC/s/mm$^2$)/(J/MAC)] energy efficiency with above mentioned implementation.

Our final benchmark focus on total energy consumption for the specific task. This kind of inference benchmarking aim to analyze how much energy is required per sample. We benchmark our PSNN architecture on the MNIST dataset, widely used in neural network performance testing. The PSNN architecture is the same as we used in benchmarking accuracy performance. Fig. 13 is the accuracy-energy benchmarking plot [59–65]. The x-axis is the performance (accuracy), and the y-axis is the MNIST sample image per energy. We can observe the correlation that higher accuracy results such as for TrueNorth (Case1) usually require more energy per image for the same target dataset. However, in our GF and future neuron version, we can achieve lower energy per image while keeping the same accuracy. Our currently developing GF neuron version can reach 1 image/nJ and even higher in future neuron versions.

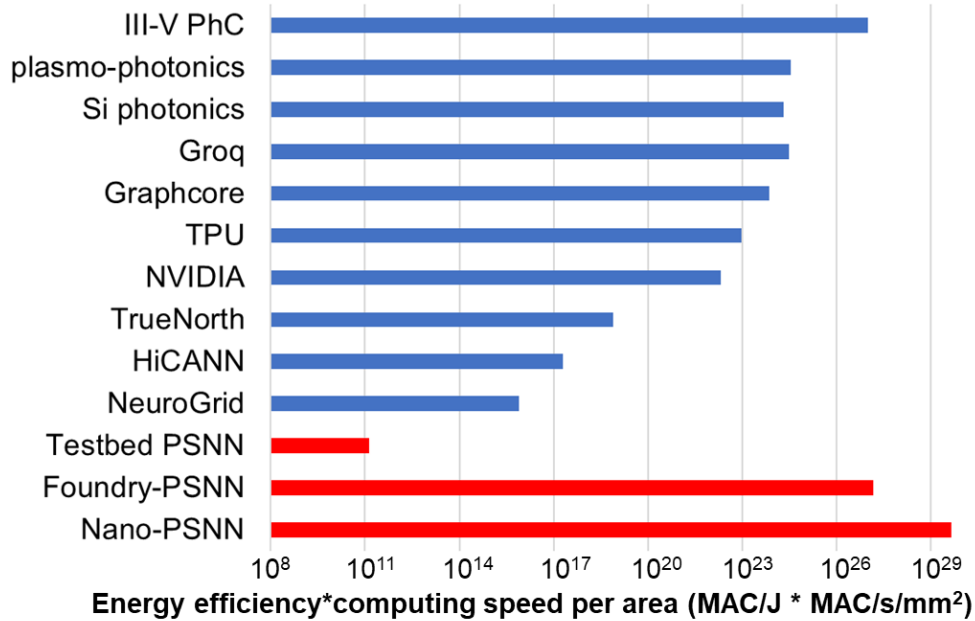

Fig. 12. Energy efficiency benchmarking based on conventional MAC operation(III-V PhC [50], plasmo-photonics [50], Si photonics [50], Groq [54], Graphcore [51], TPU [55], NVIDIA [56], TrueNorth [5] [52], HiCANN [53], NuroGrid [49])

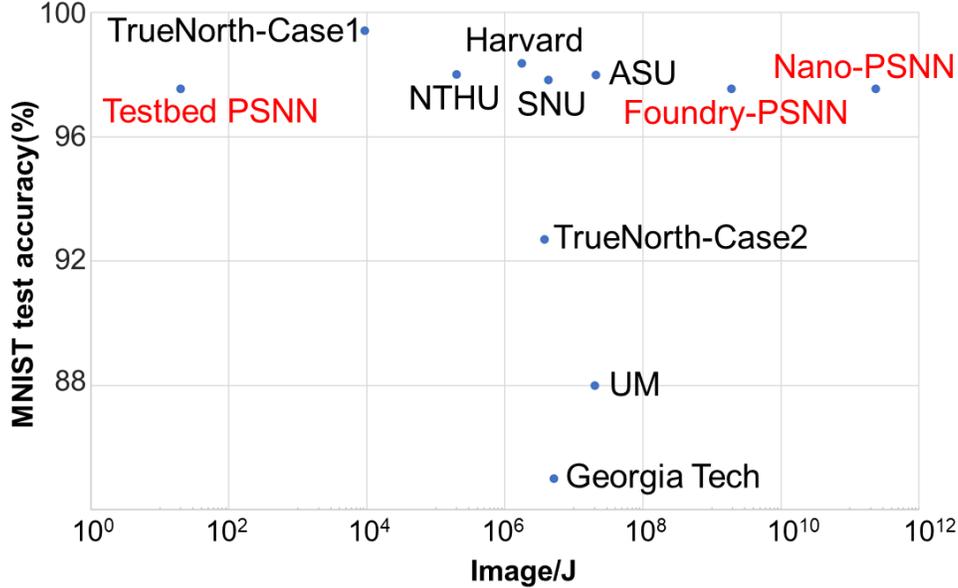

Fig. 13. Inference benchmarking targeted at MNIST image dataset (TrueNorth [5], Georgia Tech [59], Harvard [60], UM [62], SNU [63], ASU [64], and NTHU [65])

## 6. Conclusion

We designed, prototyped, and experimentally demonstrated, for the first time to our knowledge, a three transistors optoelectronic spiking neuron inspired by the Izhikevich model incorporating both excitatory and inhibitory optical spiking inputs and producing optical spiking outputs accordingly. The simulations show 97% accuracy on a supervised modified FC Nets neural network. The benchmarking result shows that PSNNs with future nanoscale optoelectronic neurons and optical-MEMS multiport interferometers can have 1000× higher energy efficiency compared to conventional electronic counterparts.

The combination of the nanoelectronic and nano-optoelectronic neuromorphic computing PSNN platforms offers new truly transformative opportunities towards realizing scalable, energy-efficient, and high-throughput edge AI systems (at your home, office, auto-vehicles, and health clinics). Instead of relying fully on Cloud computing systems, such powerful edge computing systems can handle AI tasks locally and rapidly without exposing sensitive data to the cloud and without relying on external communications at high speed. Such a powerful and energy-efficient neuromorphic computing with brain-inspired learning capabilities embedded in our everyday systems will impact many applications, including but not limited to secure and privacy-preserving IoT applications such as in smart health, autonomous vehicles, and smart cities, in particular for edge and fog computing utilizing machine learning algorithms.

## Acknowledgments

This work was supported in part by AFOSR grant FA9550-181-1-0186.
The authors would like to thank GLOBALFOUNDRIES for providing guidance for 90SIPH-90WG PDK through the 90WG university program.

# Supplementary Information for Izhikevich-Inspired Optoelectronic Neurons with Excitatory and Inhibitory Inputs for Energy-Efficient Photonic Spiking Neural Networks

## 1. Parameters Choosing Process for Izhikevich-Inspired Optoelectronic Neuron

The methodology of choosing the parameters for the Izhikevich-inspired optoelectronic neuron with excitatory and inhibitory inputs starts with building the compact model on Verilog-A, which is based on the governing equations (4)-(6) in the main paper. The simulation uses LTSpice and Cadence to determine physical parameters suitable for emulating the optoelectronic spiking neuron behaviors. All three neuron versions follow the same steps on parameter choosing. The difference between the three neuron versions is in the type of transistors, photodetectors, and the output laser. There will also be some circuit design differences between the three neuron versions, such as the number of transistors used to stabilize the voltages. Still, the neuron working mechanism remains the same. The following process will use the testbed neuron version as an example.

The process of determining parameters in the neuron circuit can be viewed in three steps. The first essential step is to decide what kinds of transistors to use on the FET1 and FET2. If we assume using the same type of transistor for FET1 and FET2, the transistor is required to support a large current on the drain terminal to the source terminal to turn VCSEL ON. Our testbed neuron demands transistors to withstand at least 10mA to excite the VCSEL. The other crucial parameter for FET1 and FET2 transistors is to determine the threshold voltage. The threshold voltage needs to be at the saturation region to perform RC charging and leaking behavior. Note that the types of FET1 and FET2 transistors are not required to be identical. However, identical FET1 and FET2 will help determine the value of the threshold.

After finding suitable FET1 and FET2 transistors, the next step is to determine the membrane potential RC circuit. The membrane potential RC circuit requires current source input, provided by the photodetector (PD1_exc) to make the circuit works. The amount of the current supplied by the photodetector is determined by the light intensity and the photodetector's responsivity. The amount of charges on each spike into the membrane potential RC circuit, the threshold voltage determined on the previous step, and the RC value will determine the neuron's charging and leaking speed. The charging and leaking speed also specify the maximum operation speed on this neuron circuit.

The final step is to determine FET3 and the refractory potential RC circuit. The capacitor value in the refractory potential RC circuit is the most crucial parameter in this step. It has to make the FET3 transistor stay in the ON state long enough to drain the membrane potential, which is the same mechanism for PD2_inh. Thus, the value of capacitor and FET3 ON state voltage need to compatible with each other. The neuron circuit-level simulation results in the main paper (Fig. 5 and Fig. 6) are based on the above methodology. The computing speed in the experiment is limited to 10 kHz for the neuron in the testbed using the available off-the-shelf transistors. The same design process applied to the Foundry PSNN and the Nano PSNN can utilize far smaller capacitor values and much faster (above 10 GHz) computing speeds.

## 2. Izhikevich-Inspired Optoelectronic Neuron Power Calculation

The power consumption of a neuron depends on the total energy required to generate certain spiking behaviors such as spikes at a certain frequency and with a certain amplitude and duration. We calculated both neuron's dynamic power and static power. The dynamic power is the power when the neuron generates spikes. The static power is the power when the neuron is resting (static off) or turning on transistors (static on) to create spikes.

The foundry neuron is simulated with a total capacitance of 68.1fF (main capacitor (60fF), photodetector load capacitance (2.1fF), and transistors parasitic capacitance (6fF)). The neuron behavior is set with three continuous spikes to charge to the threshold (0.65V). Three spikes provide a total charge of $Q = C*V = 44.27$fC, which means one spike requires to contain 14.76 fC charge. The foundry PD has a responsivity of 0.7A/W, leading to a neuron's dynamic input energy $E_{dynamic-in}$= 21.09 fJ/spike. If we assume the neuron is operating at its maximum spiking rate (10 GHz) with a spike width of 10 ps, the peak dynamic power $P_{dynamic-in}= \frac{E_{dynamic-in}}{T_{spike}}$ will be 2.11mW. To support a fanout of 10, we need an $E_{dynamic-out}$= 211fJ/spike. The output spike width is the same as the input spike width, which is 10 ps. That leads to a peak dynamic power of $P_{dynamic-out}$= 21.1mW. As for the static power in our neurons, the only power consumed is caused by the leakage current when the neuron is OFF. The neuron's leakage current is 3.18µA at 2V power supply, and 580 pA at a 0.5V power supply, making the total $P_{static-off}$= 6.36 µW. When the neuron reaches its threshold (ON), it will turn the transistors to on state. The current at 2 V power supply is 423 µA and 22.4 µA at a 0.5V power supply, leading to $P_{static-on}$= 858µW.

Assume the total neuron output spikes occupy $t$ % of the time in a certain time slot. If we assume the optoelectronic neuron is at a 10 GHz continuous spiking scenario, the neuron will turn on maximum 3.33% of the time ($t$ = 3.33). The average power is $P_{avg}= t * ( P_{dynamic-out}+ P_{static-on})+(1-t)* P_{static-off}$= 714µW. In Diehl and Cook's MNIST handwriting recognition experiment [1] mentioned in the main content, the spiking behavior is sparse. The entire neural network only spikes around 8.6% of the time in the assigned time slot. Thus, the average power for Diehl and Cook's MNIST handwriting recognition experiment will be around 61.4 µW. The total training energy will be 31.3 µJ.

The transistors reported from IRDS2020 [2] enable us to scale down on the future nano neuron because of the smaller capacitance (1.1 aF) and threshold voltage (0.1 V). Based on the working mechanism explained in section1, we can assume the future neuron will have 0.601fF of total capacitance (main capacitor (0.5 fF), photodetector load capacitance (0.1 fF) [3], and transistors parasitic capacitance (1.1 aF)). Suppose we assume the neuron behavior is the same as our previous neuron version that three continuous input spikes will charge neuron to the threshold (0.1 V), and the photonic crystal PD has a responsivity 1 A/W. In that case, we estimate the future nano neuron's dynamic input energy per spike to be $E_{dynamic-in}$= 200 aJ/spike. If we follow the same operating condition as the foundry neuron, we estimate the following values: the peak dynamic power $P_{dynamic-in} = \frac{E_{dynamic-in}}{T_{spike}}$= 20 µW, the dynamic input energy per spike $E_{dynamic-out}$= 2 fJ/spike, and the peak dynamic power $P_{dynamic-out}$ = 200 µW. The leakage current in the neuron will be 10 nA at 1.4 V power supply, making the total $P_{static-off}$= 14 nW. When the neuron is ON, the current at 1.4 V power supply is 31.27 µA, leading to $P_{static-on}$= 43.78 µW. The average power is $P_{avg}$ = 8.14µW. The power consumption of Diehl and Cook's MNIST handwriting recognition experiment will be around 0.7 µW. The total training energy to be 253 nJ. The parameters, energy values, and power values of foundry and nano neurons are listed on Table1.

Table1 Foundry and nano neurons power calculation

|  | Foundry neuron | Nano neuron |
|---|---|---|
| Maximum spiking rate | 10GHz | 10GHz |
| Spike width | 10ps | 10ps |
| C1 | 60fF | 500aF |
| PD load capacitance | 2.1fF | 100aF |
| PD responsivity | 0.7A/W | 1A/W |
| FET parasitic capacitance | 6fF | 1.1aF |
| Leakage current (voltage at power supply) | 3.18µA (2V) 580pA (0.5V) | 10nA (1.4V) |
| Neuron on current (voltage at power supply) | 423.4µA (2V) 22.4µA (0.5V) | 31.27µA (1.4V) |
| Dynamic input energy | 21.09fJ/spike | 200aJ/spike |
| Peak dynamic input power at 10GHz | 2.11mW | 20µW |
| Peak dynamic output power at 10GHz with fanout of 10 | 21.1mW | 200µW |
| Static power when neuron is on | 858µW | 43.78µW |
| Static power when neuron is off | 6.36µW | 14nW |
| Continuous spiking average power | 714µW | 8.14µW |
| MNIST handwriting recognition total training energy | 31.3µJ | 253nJ |